\documentclass[journal,twoside,web]{ieeecolor}
\usepackage{generic}
\usepackage{cite}
\usepackage{amsmath,amssymb,amsfonts}
\usepackage{algorithmic}
\usepackage{graphicx}
\usepackage{textcomp}

\usepackage{url}
\usepackage{multirow}
\usepackage{enumerate}
\usepackage[T1]{fontenc}
\usepackage{booktabs}
\usepackage{multirow}

\def\BibTeX{{\rm B\kern-.05em{\sc i\kern-.025em b}\kern-.08em
    T\kern-.1667em\lower.7ex\hbox{E}\kern-.125emX}}
\begin{document}
\title{Dynamic Interactional And Cooperative Network For Shield Machine}
\author{Dazhi Gao, Rongyang Li, Lingfeng Mao, Hongbo Wang, and Huansheng Ning, \IEEEmembership{Senior Member, IEEE}
\thanks{Corresponding author: Lingfeng Mao}
\thanks{Dazhi Gao, Rongyang Li, Hongbo Wang, Lingfeng Mao and Huansheng Ning are with the School of Computer and Communication Engineering, University of Science and Technology Beijing, 100083, Beijing, China (email: g20209484@xs.ustb.edu.cn, ustblirongyang@sina.com, foreverwhb@126.com, lingfengmao@ustb.edu.cn, ninghuansheng@ustb.edu.cn). }
\thanks{Huansheng Ning is also with Beijing Engineering Research Center for
	Cyberspace Data Analysis and Applications, Beijing, China.}
}

\maketitle

\begin{abstract}
The shield machine (SM) is a complex mechanical device used for tunneling. However, the monitoring and deciding were mainly done by artificial experience during traditional construction, which brought some limitations, such as hidden mechanical failures, human operator error, and sensor anomalies. To deal with these challenges, many scholars have studied SM intelligent methods. Most of these methods only take SM into account but do not consider the SM operating environment. So, this paper discussed the relationship among SM, geological information, and control terminals. Then, according to the relationship, models were established for the control terminal tasks, including SM rate prediction and SM anomaly detection. The experimental results show that compared with baseline models, the proposed model in this paper perform better. In the proposed model, the R2 and MSE of rate prediction can reach 92.2\%, and 0.0064 respectively. The abnormal data detection rate of the anomaly detection task is up to 98.2\%.
\end{abstract}


\section{Introduction}
\label{sec:introduction}
\begin{figure}[h]
	\centering
	\includegraphics[height=0.6\linewidth]{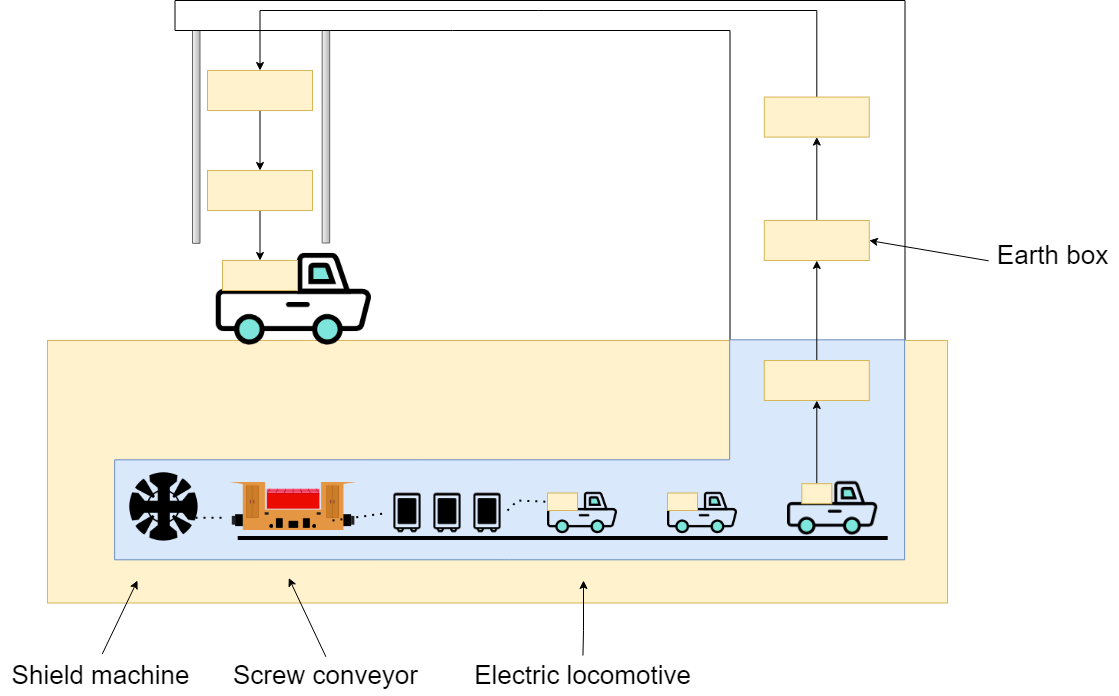}
	\caption{Shield machine operation diagram}
	\label{fig:sm}
\end{figure}

\IEEEPARstart{S}{hield} machine(SM) is a tunnel boring machine constructed by shield method. It was composed of a circular or rectangular steel cylinder structure that can support formation pressure and propel in the formation. Different from the open construction, the shield machine(SM) in the shield method uses the support structure formed by the assembly of segments to protect the tunnel during operation. This method has the advantages of fast construction speed, safety, no climate impact, little impact on the ground, and cost savings. Its operating structure diagram was shown in the figure \ref{fig:sm}.  Different from traditional cars, in actual engineering, the operation of SM requires a control terminal. When the SM is running, the control terminal will adjust the excavation parameters of the SM according to the site environment and operating conditions. After the excavation task is completed, the control terminal will also maintain the SM.

Different from traditional cars, in actual engineering, the operation of SM requires a control terminal. When the SM is running, the control terminal will adjust the excavation parameters of the SM according to the site environment and operating conditions. After the excavation task is completed, the control terminal will also maintain the SM.

In the traditional control terminal, a large number of professionals were required to participate together, and the final effect was closely related to the experience of the professional \cite{benato2015prediction}\cite{okubo2003expert}. This uncertainty also often led to higher costs and lower efficiency for construction, which seriously affected the normal progress of the project. Consequently, monitoring and decision-making mainly rely on the subjective experience of professionals in the traditional construction of SM. The inherent knowledge of data needs to be further explored and applied to improve intelligence.With the widespread use of SM, the shortcomings of the manual-based model had gradually emerged. Therefore, some scholars began to model the SM behavior by theoretical formulas to predict the SM’s behavior. Although this method had replaced part of the manual work, it still needs to be further improved.

Today, with the rise of artificial intelligence and digital twin technology\cite{xie2019virtual}\cite{Tao2019Digital}, many scholars had begun to try to use this technology to model SM and use the model for prediction and anomaly detection. At this time, the model was not only established through the theoretical model of mechanical equipment or human experience but started from the data of the whole process of the operation of the SM, to describe the SM from different angles. Although these methods made full use of the data of the SM operating period, they often ignored the role of geological information in modeling and only regarded it as a condition for division or a signal for identification. A SM was a constructing tool used below the surface, and its operating efficiency is closely related to geological conditions. Many changes in the excavation parameters of the SM during excavation were caused by geological changes, and some sudden changes in excavation parameters were often caused by sudden changes in geology. Therefore, geological information were important parameters that cannot be ignored in the SM modeling. If the geological information is ignored, it will be difficult for the model to make correct responses when the geological information suddenly changes, reducing the accuracy and robustness of the model.

In this paper, the relationship among SM, geological information, and control terminals was combed in detail and discussed the significance of geological information and excavation parameters in the task of SM. 
According to this, two classical tasks of the control terminal are proposed, rate prediction of the SM and SM abnormal detection are more suitable for the SM method.

The main contributions of this paper are as follows:

1. In order to solve the problem of poor model performance and robustness caused by ignoring geological parameters in SM modeling, demonstrate the synergistic and crossfeed relationship among SM, geological information, and control terminal in SM modeling, and propose the importance of geological information in SM modeling

2. To solve the problem of SM rate prediction of the control terminal, a Convolutional Neural Network(CNN)\cite{lecun1989backpropagation} model based on the attention mechanism is proposed. Compared with the baseline model, the performance is greatly improved, and the influence of adding geological information to the model is compared.

3. To solve the problem of abnormal control terminal SM, an anomaly detection model based on Variational Auto-Encoder(VAE) and Long Short-Term Memory(LSTM)\cite{graves2012long} is proposed, which improves the accuracy of anomaly recognition compared to the baseline model.

The remainder of this paper was organized as follows. The section II reviewed the development of shield anomaly detection and tunneling parameter prediction in recent years. The third section analyzes the synergy and dynamic interactional relationship among geological information, control terminal and SM. Based on the analysis in Section III, the task model of rate prediction of the SM and anomaly detection of the SM is established in Section IV. In order to verify the effectiveness of the model, in Section V we present the characteristics of the data and the method of data preprocessing. In Section VI, this paper compares the performance of the two proposed models with baseline models respectively.

\section{Related Works}\label{METHODS BASED ON SHORT DATA}

\subsection{Rate Prediction of the SM}\label{A.	Data change in short data}

The research on the rate prediction of the SM began as early as the 20th century and could be divided into experience-based models, statistics-based models, and machine learning-based models.

Experience-based models used relevant knowledge of physics to physically simulate the SM to predict future changes. Wang et al.\cite{wang1978tunnel} studied the theoretical equation of tunnel drillability. Benato et al.\cite{benato2015prediction} established a prediction model for the rock penetration rate in Tunnel Boring Machine(TBM) operation through previous empirical models, which were used to calculate the penetration rate each time, and successfully for instance. Hassanpour et al.\cite{hassanpour2010tbm} also analyzed various parameters of TBM and compared them with the predictive model, and finally obtained the better method and used it to establish an empirical model. Rostami et al.\cite{rostami1997development} used an empirical model to predict cutting forces on a disc milling cutter. Li et al.\cite{li2021advanced} used the data processing group method to predict the penetration rate of TBM, and the effect was better than five multiple regression models. Most of the experience-based models were only designed for one area and one problem and did not have high generalization.

With the continuous development of machine learning, some scholars also tried to use machine learning methods to solve prediction problems. Ghasemi et al.\cite{ghasemi2014predicting} used fuzzy logic to predict the drilling rate of rock properties such as uniaxial compressive strength, rock brittleness, the distance between weak planes, and direction of discontinuities in the rock mass and obtained good results. Benardos et al.\cite{benardos2004modelling} used artificial neural network to predict the advancing speed of the excavation site. Armaghani et al.\cite{armaghani2017development} used two algorithms, particle swarm optimization (PSO) artificial neural network (ANN) and imperialist competition algorithm to predict TBM penetration rate. Koopialipoor et al.\cite{koopialipoor2019application} used Deep Neural Network (DNN) to build a network for predicting TBM penetration and compared it with ANN, verifying the performance superiority of DNN compared to ANN. Mahdevari et al.\cite{mahdevari2014support} used support vector regression(SVR) to predict penetration rate. Fattahi et al.\cite{fattahi2017applying} used SVR and differential evolution algorithm (DE)\cite{price2013differential}, artificial bee colony algorithm (ABC) and gravity search algorithm (GSA) to comprehensively predict TBM performance.

With the continuous development of deep learning, the use of deep learning models to solve problems has become a trend today. Li et al.\cite{li2021advanced} used LSTM to predict the total thrust and cutter head torque in the stable period with the data of the rising period of the TBM tunneling section, and the performance was better than the traditional empirical model. Nagrecha et al.\cite{nagrecha2020encountered} used recurrent networks(RNN)\cite{elman1991distributed} to predict the parameters of TBM.

\subsection{Anomaly Detection of the SM}\label{Feature space in short data}
anomaly detection of the SM started early, and the model had a wide range of applications\cite{9635688}. However, it also faced the problem of less negative data and a lack of expert datasets.

Expert rule-based methods were often used for early detection. Okubo et al.\cite{okubo2003expert} created an expert system for tunnel boring machines and applied them to Japanese tunnel boring machines. Sutherland et al.\cite{sutherland2000performance} believed that the performance of the tunnel boring machine depends on the mine power distribution system. So the author proposed to use the physical formula to model the motor and compare it with the actual voltage. Chen et al.\cite{chen2018building} established a shield construction data analysis platform, and used association rules to implement automatic motor current anomaly detection rules, which were applied to this platform.

Today, regression and clustering-based methods were more commonly used in anomaly detection. Sheil et al.\cite{sheil2020assessment} used the methods based on clustering and regression to detect abnormal jacking forces in the process of micro-advancement and made their own evaluation of the two methods. Grima et al.\cite{grima2000modeling} proposed the neuro-fuzzy method to model the tunnel boring machine, which pioneered the use of fuzzy theory combined with neural networks to solve geological engineering problems. Bai et al.\cite{bai2021pipejacking} evaluated the potential of One-class support vector machine(OCSVM), Isolation Forest, and Robust Covariance in this problem for the detection of clogging problems often encountered in clay tunnelling engineering, and received good responses. Sipers et al.\cite{sipers2017robust} developed unthresholded recurrence plot (URP) to achieve dynamic representation of features, and then introduced extreme learning machine auto-encoder (ELM-AE) to detect abnormal data.

\section{Relationship analysis of shield machine, geological information, and control terminal}\label{Methods dealing with missing data}

\begin{figure}[h]
	\centering
	\includegraphics[height=0.6\linewidth]{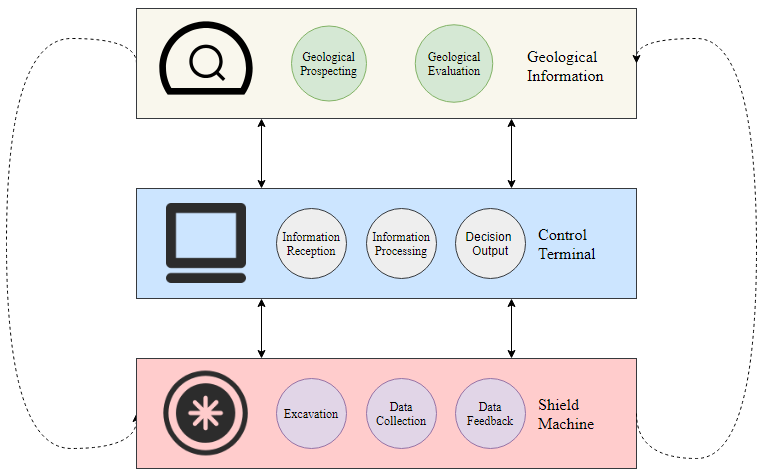}
	\caption{The relationship among  geological information, control terminal and shield machine}
	\label{fig:relation}
\end{figure}

The tunneling work of the SM was driven by the rotation of the cutter head and the propelling cylinder. With the continuous advancement of the SM, the soil cut by the cutter head would be put into the soil bin. When the soil bin was filled with soil, the screw conveyor will be opened to discharge the soil to the belt conveyor, and finally, it would be put into the soil box of the electric locomotive and transported to the ground by gantry hoisting. From this, we could conclude that the operation of the SM was the result of the cooperative work of various parts inside and outside the SM. From the overall point of view, it was the result of the collaboration between the SM and the geological information. Therefore, there was an inseparable synergistic relationship between the SM, geology, and control terminals, which together promote the smooth operation of the SM. The relationship among the three was shown in the figure \ref{fig:relation}.

\subsection{ The dynamic interaction between shield machine and control terminal}\label{Statistics-based method in missing data}

The reason why the control terminal was originally created was to maintain the smooth operation of the SM. Therefore, there was a direct dynamic interactional relationship between the SM and the control terminal. First of all, each key component of the modern SM was equipped with data collection equipment, which collects real-time operating parameters at certain time intervals and transmits them to the data collection terminal. After the data collection terminal collects these data, it would be sent to the cloud server, and finally sent to the control terminal. The control terminal obtained the operation status and health status of the SM according to the changes in the real-time operating parameters, and then judged whether it needs adjustment and maintenance. Finally, the parameters were adjusted according to the judgment result. The control terminal receives the operating parameters of the SM to make judgments, and then feedback the adjustment situation to the SM, and then the SM dynamic interactional the operating conditions under this parameter to the control terminal. A good mutual dynamic interactional relationship ensured the stable operation of the SM.

\subsection{The dynamic interaction between geological information and control terminals}\label{Machine learning in missing data}

The geological information and the control terminal form a special dynamic interactional relationship. First of all, before each shield excavation task started, professional geological prospectors were required to conduct detailed geological surveys on the road sections to be excavated and handed over to the SM operators. The personnel formulated the construction plan of this route according to the geological information obtained from the survey. During the excavation task, professionals will also adjust at any time according to the geological information. Therefore, an indirect dynamic interactional relationship is formed between the geological information and the control terminal.

\subsection{The collaboration between geological information and shield machine}\label{Machine learning in missing data}

The relationship between the geological information and the SM was not direct physical coordination. The geological information and the SM do not communicate information directly but use the control terminal as a intermediation. For the control terminal, the data used for analysis was incomplete if it was only shielded information or just geological information. The excavation information was the result of the interaction between the SM and the geological information. Likewise, the analysis of geological information alone was not sufficient for the overall excavation task. The control terminal analysis data was not a kind of geological information and shield information, but a fusion analysis of the two pieces of information, so as to obtain relatively objective results that reflect the actual situation. Therefore, a collaborative relationship with the control terminal as the medium was formed between the geological information and the SM.

\section{Collaborative and dynamic interactional network application construction}\label{Methods based on short label}

The SM collaborative dynamic interactive network application was built in the control terminal. The control terminal processed the received geological information and SM information, and then feed back the processing results. This paper focused on two classic tasks for shield machines: SM rate prediction and SM anomaly detection,  proposed two models based on Collaborative and dynamic interactive network.

In order to provide data support for model validation, this paper used the dataset from 1 to 400 rings in a real section, and the characteristics of the geological information are as follows Table \ref{geo}.The excavation data of the SM contained the following characteristics as shown in Table \ref{tun}.

As can be seen from Table \ref{geo} and Table \ref{tun}, the data had the characteristics of different types, large dimension gaps and many features, and cannot be processed directly. Therefore, the preprocessing method in this task would be given in the next section.

\subsection{Rate prediction of the shield machine construction}

\begin{figure}
	\centering
	\includegraphics[width=0.7\linewidth]{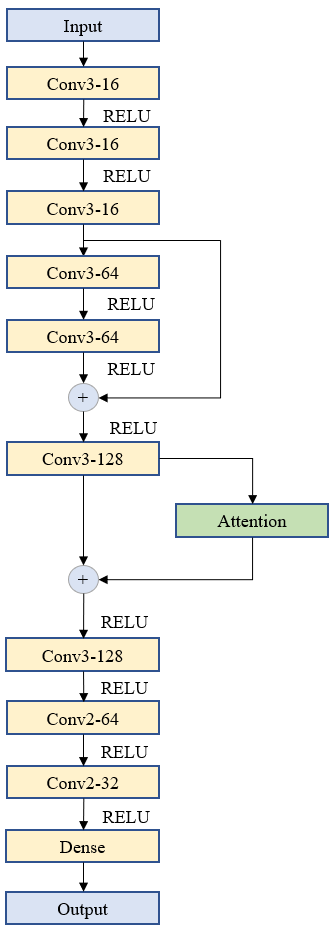}
	\caption{Rate prediction of the shield machine model structure}
	\label{fig:screenshot001}
\end{figure}

In order to ensure that the SM can adapt to different geological conditions, there were a large number of parameters that could be adjusted according to the site conditions during the excavation process of the SM. The SM rate is a very important parameter that reflects the operation of the shield machine. The SM rate is a parameter for evaluating the operating speed of the shield machine at a certain time, and its change is very sensitive. The factors that cause the change can be summarized as human factors, the shield machine itself and geological factors. In the previous prediction models, only human factors and the shield machine itself are often concerned, and geological factors are added only in the result evaluation. Based on the collaborative and dynamic interactional network proposed above, this paper proposes a CNN model based on the attention mechanism that adds geological information. The model shown in Figure \ref{fig:screenshot001}.

This paper used modules such as CNN, attention, residual network, linear layer and dropout in this model, and the activation function adopted RELU. Among them, CNN, as a powerful feature extractor, could well extract the features of the input data. The participation of the attention mechanism could assign corresponding weights to different features, allowed the model to invest more energy in more important features. Residual connections and dropout layers can alleviated model overfitting.  At the same time, the loss function of this network adopted the $Smooth L_{1}$ function, and its formula as follows:
\begin{equation}
\operatorname{Smooth L_{1}}(x)=\left\{\begin{array}{ll}
	0.5 x^{2} &  { if } |x|<1 \\
	|x|-0.5 &  { otherwise }
\end{array}\right.
\label{smo}\end{equation}

$Smooth L_{1}$ had higher robustness than L1 and L2 loss functions, and was insensitive to outliers, had better results in this task.

\subsection{Anomaly detection of the shield machine construction}
\begin{figure}
	\centering
	\includegraphics[width=0.7\linewidth]{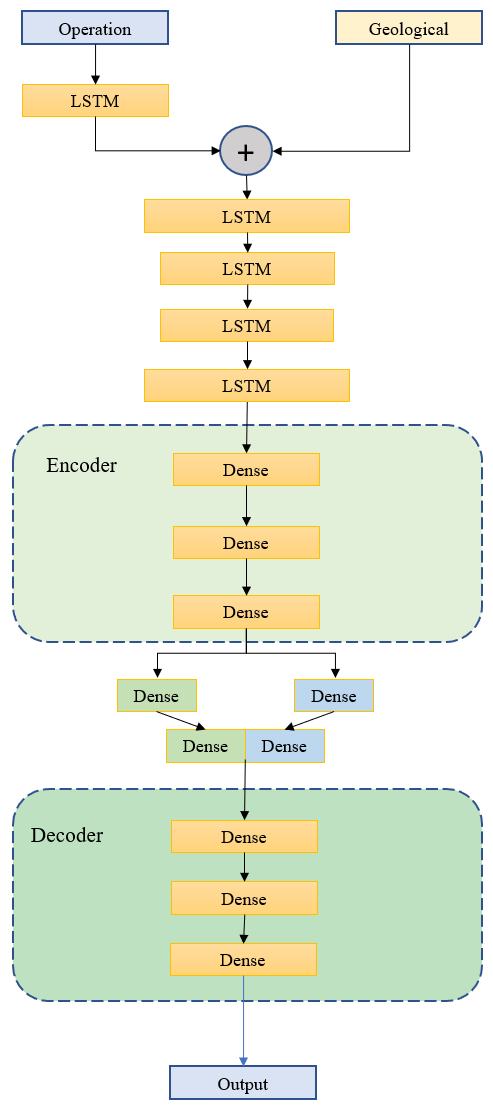}
	\caption{Shield machine anomaly detection model structure}
	\label{fig:ab}
\end{figure}

SM anomaly detection is a method to detect the operation of the shield machine when the SM is running, which is generally obtained by analyzing the output data of the SM. The occurrence of shield machine anomalies has certain randomness and environmental correlation, so it is necessary to add geological data in the analysis to obtain reasonable results. This paper also proposed a SM anomaly detection method based on LSTM and Variational Autoencoder(VAE)\cite{an2015variational} model based on the collaborative and dynamic interactional network. The model structure diagram can be shown in \ref{fig:ab}.

The VAE model was a generative model developed from the Auto Encoder(AE). The AE model consists of an encoder $f(X)$ and a decoder $g(X)$. The encoder encodes the input sample  $X=\left\{x_{1}, x_{2}, \cdots \cdots x_{n}\right\}$ to obtain a latent vector $Z$, and then the decoder obtains the generated value $X^{\prime}=\left\{x_{1}^{\prime}, x_{2}^{\prime}, \cdots \cdots x_{n}^{\prime}\right\}$ .  The purposed of the AE model was to make $X^{\prime}$ as close to $X$ as possible, so that we can learn $Z$ that contains model information. In SM anomaly detection, when the $Z$ learned by using the information of normal road sections encounters abnormal information, because the abnormal information did not conform to the distribution of normal information, the generated result would have an abnormal error with the input, we could follow this determine whether there was an abnormality. However, in the AE model, the encoded $Z$ is not modeled, and the uncertainty of the $Z$ distribution also made the sampling had a certain randomness, resulting in a certain randomness of the generated effect.

In order to solve the uncertainty of the $z$ distribution, the VAE model used $q(Z \mid X)$ to approximate the conditional probability distribution $p(Z \mid X)$ of $Z$ and $X$. For each real sample $x_{k}$,
determine the exclusive conditional probability distribution $p(z \mid x_{k})$, 
and obtained the mean vector $\mu_{k}$ and variance vector $\sigma^{2}$ 
in the normal distribution of each sample $x_{k}$, 
from which the exclusive distribution of each real sample can be obtained, and thus obtain $Z$ by sampling $z_{k}$.
To measure the similarity between distributions, KL divergence was added to the loss function. Its formula referred to \eqref{kl}:
{\small
	\begin{equation}
		\operatorname{loss}_{\mathrm{kl}}=K L(P(z \mid x) \| q(z \mid x)) \\
	\label{kl}\end{equation}
}
{\small
	\begin{equation}
		K L(P(z \mid x) \| q(z \mid x))=\frac{1}{2}\left(-\log \sigma^{2}+\mu^{2}+\sigma^{2}-1\right)
		\label{kl}\end{equation}
}

where $N$ was the number of samples, $w_{k}$ was the weight.  Therefore, the loss function used in this model as follows:

\begin{equation}\operatorname{loss}_{B C E} \cdot=\frac{1}{N} \sum_{n=1}^{N} l_{n}\label{l1}\end{equation}
\begin{equation}l_{n}=-w_{k}\left[x_{k}^{\prime} \log x_{k}+\left(1-x_{k}^{\prime}\right) \log \left(1-x_{k}\right)\right]\label{l2}\end{equation}
\begin{equation}\operatorname{loss}=\operatorname{loss}_{k l}+\operatorname{loss}_{B C E}
\label{l3}\end{equation}

LSTM was a network for processing time series data, developed from the RNN model. LSTM added a gating mechanism to alleviate the problems of long-term dependencies, gradient disappearance and gradient explosion in the RNN model that cannot be memorized. The LSTM model had a better effect on the high-dimensional and long-term data of the SM. This model was a two-head input model, and the features of the two inputs could be well fused with the help of the characteristics of LSTM. Therefore, the LSTM layer was pre-trained by the AE model before all model training. The purpose of this training was to allow the LSTM layer to better integrate the double-headed input.

\begin{table}[]
	\centering 
	\caption{Geological types and their samples}
	\label{geo}
	\setlength{\tabcolsep}{10mm}{
	\begin{tabular}{@{}ll@{}}
		\toprule
		Geological Features           & Example             \\ \midrule
		Plasticity                    & Soft plastic \\
		Density                       & Loose               \\
		Unconfined Compressive Strength & 27.8                \\
		Permeability Coefficient       & 0.0115740740740741  \\
		Surrounding Roc Level           & 5                  \\
		Layer Number                   & 3         \\
		Accounting                    & 0.0087              \\
		Integrity Factor               & 0.36 to 0.51           \\
		Standard Penetration           & 9.8                 \\ \bottomrule
	\end{tabular}}
\end{table}

\begin{table}[]
	\centering 
	\caption{Excavation data and their samples}
	\label{tun}
	\setlength{\tabcolsep}{10mm}{
	\begin{tabular}{@{}ll@{}}
		\toprule
		Shield Tunneling Features & Example  \\ \midrule
		Propulsion Speed          & 3.82     \\
		Cutter Speed              & 1.5      \\
		Cutter Torque             & 2080.6   \\
		Total Propulsion          & 40784.81 \\
		Cutter power              & 322      \\
		Displacement              & 680      \\
		Propulsion Pressure       & 65.93    \\
		Propulsion Thrust         & 9488.69  \\ \bottomrule
	\end{tabular}}
\end{table}

\section{Data preprocessing}

\subsection{Rate prediction of the shield machine}
In the rate prediction of the SM task, the input of data was geological information and tunneling data.

(1) Obtain geological information. Before the excavation task starts, the detailed geological information of the area needed to be obtained, so this model also needed to first obtain the geological data before using it, and used the Word2Vec\cite{goldberg2014word2vec} model for the word vectorization of the text part.

(2) Obtain historical excavation data and eliminate other variables unrelated to excavation tasks. For example, "Dug Mode" and "Assembly Mode", only the variables related to the excavation task are retained.

(3) Data merging. According to the progress of the excavation, the geological information and the excavation data were combined to generate the geology- excavation data.

(4) Data completion and deletion of discrete points. After the geological-excavation data was obtained, the missing data was completed and discrete points are performed, wherein the discrete points were values that greatly deviate from the normal time series changes.

(5) Data normalization. In order to eliminate the influence of different dimensions between different data on the results, the data were normalized, the formula referred to \eqref{datafr}.
\begin{equation}
X^{\prime}=\frac{X-X_{mean}}{X_{std}}
.\label{datafr}\end{equation}
where $X^{\prime}$ represented the normalization result, $X$ was the original sequence to be normalized, $X_{mean}$ referred to the mean value of the original sequence, and $X_{std}$ represented the standard deviation of the original sequence.

(6) Delete the rising period and non-running period data. Select the data in the ascension period, assembly and maintenance period of the SM to delete.

\subsection{Anomaly detection of the shield machine}
In the abnormal data detection of the SM task, the input of the model also included geological information and excavation data.

If it was an input time series, the data preprocessing contains the following steps:

(1) Extract the operating state of the device. Only the parameters of the input running state are saved, and missing values are filled.

(2) Remove outliers from the sequence data, use the window average method for smoothing, and used the box-cox\cite{sakia1992box} test to eliminate part of the noise of the data.

(3)The data was normalized by the method of normalizing the maximum and the minimum value. Equation \eqref{data} shows the formula used for normalization.
\begin{equation}X^{\prime}=\frac{X-X_{min }}{X_{max }-X_{min }}.\label{data}\end{equation}
Where $X^{\prime}$ was the normalized result, $X$ referred to the original sequence, $X_{max }$ and $X_{min}$ were the maximum and minimum values of the original sequence.

For geological information, this paper only needed to make it match the input text according to the time axis, so that the running parameters could match the geological information.

\section{Models Validation and Analysis}
In this section, the previously established model would be verified using real data sets and the effectiveness of the model would be analyzed.
\subsection{Shield machine rate prediction model validation and analysis}
In the rate prediction of the SM task, we divided the collected real operation data into training set, validation set and test set according to the ratio of 7:2:1. The training set and validation set were used for model training and auxiliary training. We used $R_{2}$ and MSE to evaluate the performance of the model, and its calculation formula referred to \eqref{MSE} \eqref{r2}.
\begin{equation}
	\begin{aligned}
		M S E &=\frac{1}{N} \sum_{i=1}^{N}\left(y_{i}-\hat{y}_{i}\right)^{2} \\
	\end{aligned}.\label{MSE}\end{equation}

\begin{equation}
	\begin{aligned}	
		R^{2} &=1-\frac{\sum_{i=1}^{N}\left(y_{i}-\hat{y}_{i}\right)^{2}}{\sum_{i=1}^{N}\left(y_{i}-y_{m}\right)^{2}}
	\end{aligned}.\label{r2}\end{equation}

where $N$ was the number of samples, $y_{i}$ referred to the real value, $\hat{y}_{i}$ represented the predicted value obtained by the model, and $y_{m}$ was the average value of the real value.

Table \ref{adjust} gave a comparison of the model's performance on whether to add geological information, and whether to use residual network and attention under the same data. It could be seen from the table that using geological information could allow the model to learn more features and have a better fit to the data. This paper believe that the reason for this phenomenon was that geological information can make the changes of the excavation data more basis, and make the model understand more "thoroughly". At the same time, the excavation data had high-dimensional and long-distance features, so a deeper model was required to learn the information. The use of residual connections could make the information transfer of the model more smooth, reduce the information difference caused by the depth of the model, and alleviate the problem of gradient dispersion. The attention mechanism could help the model select the importance of features and make the model learning more targeted.

\begin{table*}[]
	\centering 
	\caption{Rate prediction of the shield machine model structure comparison}
	\label{adjust}
	\setlength{\tabcolsep}{10mm}{
	\begin{tabular}{@{}cccc@{}}
		\toprule
		Whether to add geological information                & modules used               & $R^{2}$     & MSE    \\ \midrule
		\multirow{4}{*}{Geological data included}     & Use attention and residual & \textbf{92.2\%} & 0.0065 \\ \cmidrule(l){2-4} 
		& Use attention              & 89.8\% & \textbf{0.0064} \\ \cmidrule(l){2-4} 
		& Use residual               & 91.8\% & 0.0065 \\ \cmidrule(l){2-4} 
		& Only use CNN               & 91.5\% & 0.0067 \\ \midrule
		\multirow{4}{*}{Geological data not included} & Use attention and residual & 91.9\% &   0.0065     \\ \cmidrule(l){2-4} 
		& Use attention              & 87.2\% &   0.0067     \\ \cmidrule(l){2-4} 
		& Use residual               & 91.4\%       & 0.0069       \\ \cmidrule(l){2-4} 
		& Only use CNN               & 91.3\% & 0.0069 \\ \bottomrule
	\end{tabular}}
\end{table*}

To demonstrate the superiority of this model, this paper compared this model with LSTM, RNN, TCN, CNN and CNN-LSTM used the same data, as shown in Table \ref{compare1}. It could be seen that the RNN and LSTM models cannot learn the characteristics of the data well and had lower $R^{2}$ and MSE values. The CNN and CNN+LSTM models were better than the above two models, but the learning of dataset features was still incomplete.

\begin{table}[]
	\centering 
	\caption{Rate prediction of the shield machine model compared}
	\label{compare1}
	\setlength{\tabcolsep}{10mm}{
	\begin{tabular}{@{}cccc@{}}
		\toprule
		Method   & $R^{2}$      & MSE \\ \midrule
		LSTM     & 89.6\% & 0.1471   \\
		RNN      & 87.7\% & 0.1508   \\
		CNN      & 91.2\% & 0.1499    \\
		CNN+LSTM & 91.3\% & 0.1489    \\
		TCN & 88.35\% & 0.1552    \\
		Proposed & \textbf{92.2\%} &  \textbf{0.0065}   \\ \bottomrule
	\end{tabular}}
\end{table}

\subsection{Shield machine anomaly detection model validation and analysis}
In this task, the data used includes normal segment data and abnormal segment data. The normal segment was the data generated during normal operation, and both were normal data. The abnormal segment contains 114 abnormal data information marked out. This task used normal segment data during training, and used abnormal segment data during testing and evaluation. This paper argued that in the anomaly detection, the purposed model was not to reconstruct the input data, but to learn the data distribution. Therefore, this paper would only used the abnormal data detection rate to verify the model effect, and its formula referred to \eqref{acc}.
\begin{equation}a c c=\frac{x_{all}-x_{pre}}{x_{all}}.\label{acc}\end{equation}
where $x_{all}$ was the labeled outlier, and $x_{pre}$ was the outlier predicted by the model.

This model used the LSTM model to fuse the input excavation data and geological information, which increased the richness of the input data and bringed richness to the data changes. Some normal changes provided a basis, and the identification of abnormal changes was also improved, which improved the effect of anomaly detection.

To demonstrate the superiority of this scheme, this paper compared this model with AE, isolation forest and K-Nearest-Neighbours (KNN)\cite{arroyo2009forecasting} models, as shown in the table \ref{compare2}. It could be seen that for the complex data of the SM. After comparison, this model had the best performance in the anomaly detection of the SM task.

\begin{table}[]
	\centering 
	\caption{Anomaly detection of the shield machine model compared}
	\label{compare2}
	\setlength{\tabcolsep}{7mm}{
	\begin{tabular}{@{}ccc@{}}
		\toprule
		Module           & Accuracy        \\ \midrule
		AE               & 68\%            \\ \midrule
		Isolation forest & 62\%            \\ \midrule
		KNN              & 83\%            \\ \midrule
		Proposed         & \textbf{98.2\%} \\ \bottomrule
	\end{tabular}}
\end{table}

\section{Conclusion}
In this paper, the relationship between SM, control terminals, and geological information was analyzed. It was considered that geological information plays a very important role in the control operation of the SM, which can reflect the correct state of the SM operation, rather than just based on what the SM parameters reflect. This paper discussed the relationship between the three in detail and discovery the coordinated and dynamic interactional relationship among the control terminal, geological, and excavation parameters. Afterward, to verify this relationship, we built models for two important tasks in the control terminal: rate prediction of the SM and abnormal data detection of the SM, respectively, and verify them with real operating data. After comparison, the two models proposed in this paper reaching the better performance. This paper believe that this effect was inseparable from the participation of geological information. Because geological information was an important determinant of driving parameters in real operation, many changes in driving parameters were often caused by changes in geological information. So introducing geological information may be a good way to improve the model effect.

\bibliographystyle{IEEEtran}
\bibliography{referenc1e}

\begin{IEEEbiography}[{\includegraphics[width=1in,height=1.25in,clip,keepaspectratio]{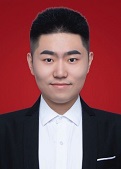}}]{Dazhi Gao} received the B.S degree from Shandong Jiaotong University, China, in 2020. He is currently working toward the Master degree from the School of Computer and Communication Engineering, University of Science and Technology Beijing, China. His research interests include Internet of Things and Artificial Intelligence.
\end{IEEEbiography}

\begin{IEEEbiography}[{\includegraphics[width=1in,height=1.25in,clip,keepaspectratio]{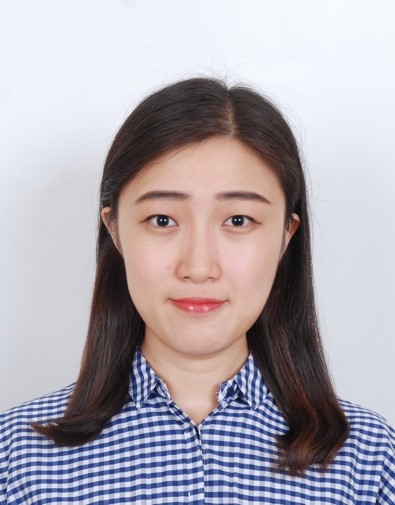}}]{Rongyang Li}  received her B.S. degree from Hebei Normal University of Science and Technology in 2018 and her M.S. degree from University of Science and Technology Beijing in 2021. She currently working toward her Ph.D. degree in the School of Computer and Communication Engineering, University of Science and Technology Beijing, China. Her current research focuses on the user experience of games.
\end{IEEEbiography}

\begin{IEEEbiography}[{\includegraphics[width=1in,height=1.25in,clip,keepaspectratio]{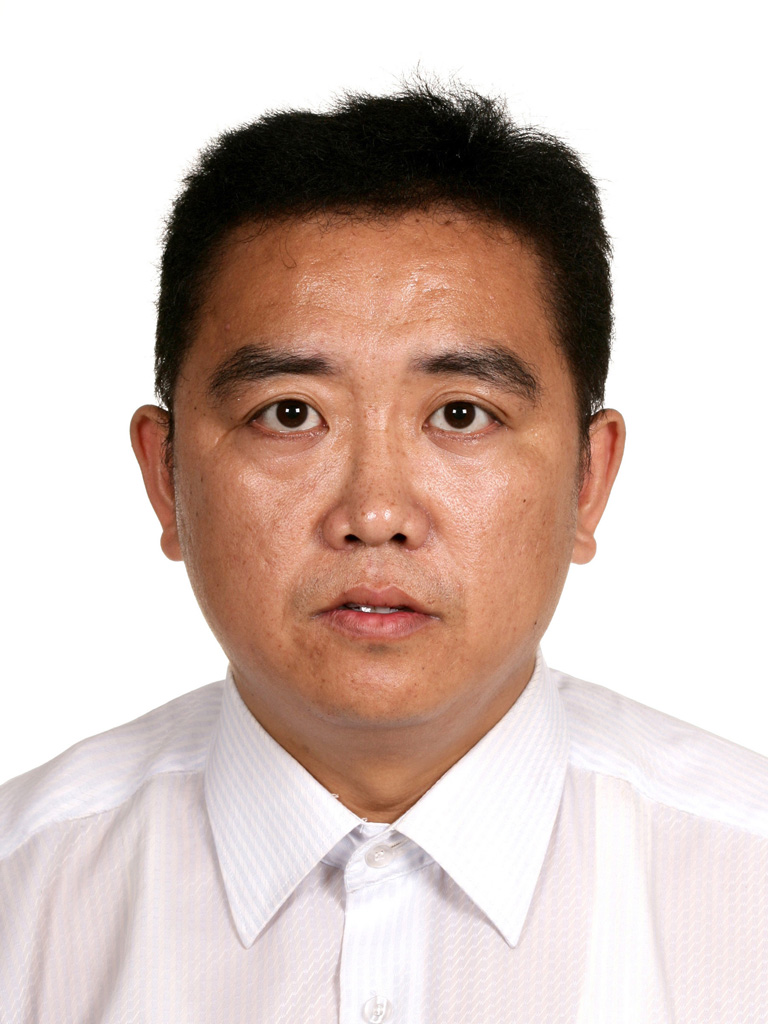}}]{Hongbo Wang}  received the doctor degree in computer application technology from University of
	Science and Technology Beijing,
	China, in 2007. After graduation,
	he joined the School of Computer
	and Communication Engineering,
	USTB, where he is an Associate
	Professor and M.S. candidate supervisor, his main research interests include natural computing algorithm,
	intelligence optimization scheduling.
\end{IEEEbiography}

\begin{IEEEbiography}[{\includegraphics[width=1in,height=1.25in,clip,keepaspectratio]{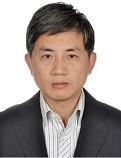}}]{Lingfeng Mao}  received the Ph.D. degree from
	Peking University, Beijing, China, in 2001.
	He is currently a Professor with the School
	of Computer and Communication Engineering,
	University of Science and Technology Beijing,
	Beijing, and the Beijing Engineering Research
	Center for Cyberspace Data Analysis and
	Applications, Beijing. His current research interests
	include the modeling and characterization of
	semiconductors, semiconductor devices, integrated
	optic devices, and microwave devices.
\end{IEEEbiography}

\begin{IEEEbiography}[{\includegraphics[width=1in,height=1.25in,clip,keepaspectratio]{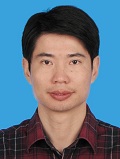}}]{Huansheng Ning}  received his B.S. degree from Anhui University in 1996 and his Ph.D. degree from Beihang University in 2001. Now, he is a professor and vice dean of the School of Computer and Communication Engineering, University of Science and Technology Beijing, China. His current research focuses on the Internet of Things and general cyberspace. He is the founder of the Cyberspace and Cybermatics International Science and Technology Cooperation Base.
\end{IEEEbiography}

\end{document}